\documentclass{article}

\usepackage[preprint,nonatbib]{neurips_2023}

\usepackage[utf8]{inputenc} 
\usepackage[T1]{fontenc}    
\usepackage{hyperref}       
\usepackage{url}            
\usepackage{booktabs}       
\usepackage{amsfonts}       
\usepackage{nicefrac}       
\usepackage{microtype}      
\usepackage{xcolor}         

\usepackage[acronym]{glossaries}
\usepackage{graphicx}
\usepackage{amsmath}
\usepackage{booktabs}
\usepackage{multirow}
\usepackage{amssymb}
\usepackage{url}
\usepackage{subcaption}
\usepackage{tabularx}
\newcolumntype{C}{>{\centering\arraybackslash}X}

\title{NeMO: Neural Map Growing System for Spatiotemporal Fusion in Bird's-Eye-View \\ and BDD-Map Benchmark}

\author{%
	Xi Zhu$^*$\quad Xiya Cao\thanks{Equal contribution.} \quad Zhiwei Dong\quad Caifa Zhou\quad
	Qiangbo Liu\quad Wei Li\quad Yongliang Wang\thanks{Corresponding author. Email: wangyongliang775@huawei.com} \\
	Riemann Lab, 2012 Laboratory, \\
	Huawei Technologies Co. Ltd\\
}

\begin{document}
\newacronym{bev}{BEV}{Bird's-Eye-View}
\newacronym{pv}{PV}{Perspective View}
\newacronym{ads}{ADS}{Autonomous Driving System}
\newacronym{rnn}{RNN}{Recurrent Neural Network}
\newacronym{lstm}{LSTM}{Long Short-Term Memory}
\newacronym{mlp}{MLP}{Multi-layer Perceptron}
\newacronym{miou}{mIoU}{mean intersection-over-union}
\newacronym{lsa}{LSA}{local-spatial attention}

\maketitle

\begin{abstract}
Vision-centric Bird’s-Eye View (BEV) representation is essential for autonomous driving systems (ADS). Multi-frame temporal fusion which leverages historical information has been demonstrated to provide more comprehensive perception results. While most research focuses on ego-centric maps of fixed settings, long-range local map generation remains less explored. This work outlines a new paradigm, named NeMO, for generating local maps through the utilization of a readable and writable big map, a learning-based fusion module, and an interaction mechanism between the two. With an assumption that the feature distribution of all BEV grids follows an identical pattern, we adopt a shared-weight neural network for all grids to update the big map. This paradigm supports the fusion of longer time series and the generation of long-range BEV local maps. Furthermore, we release BDD-Map, a BDD100K-based dataset incorporating map element annotations, including lane lines, boundaries, and pedestrian crossing. Experiments on the NuScenes and BDD-Map datasets demonstrate that NeMO outperforms state-of-the-art map segmentation methods. We also provide a new scene-level BEV map evaluation setting along with the corresponding baseline for a more comprehensive comparison.	

\end{abstract}

\section{Introduction}

In the realm of autonomous driving, the ability to perceive and comprehend the surrounding environment is of utmost importance. The \acrfull{bev} representation is particularly desirable for its ability to accurately display the spatial placement of objects and road elements in a three-dimensional space \cite{wang2022detr3d,liu2022petr,can2021structured,jiang2022polarformer,zhou2022cross,bartoccioni2023lara}. Many vision-based \acrfull{bev} studies perception \cite{ammar2019geometric,li2022hdmapnet,reiher2020sim2real,hu2021fiery,peng2023bevsegformer,guo2020gen} have shown significant progress in recent years. 

Besides utilizing multi-view information\cite{li2022hdmapnet, hu2021fiery}, vision-based \acrfull{bev} perception also taps into the potential of time-series images. Leveraging time-series data can effectively address challenges such as visual occlusion and visual illusions, particularly for static elements, like road elements. Researchers have explored temporal fusion via warping or query-based approaches\cite{huang2022bevdet4d, liu2022petrv2, qin2022uniformer, zhang2022beverse, li2022bevformer}, demonstrating that temporal fusion improves perception in challenging environments such as occluded circumstances. However, these works limit to ego-centric environment and do not consider the global perception of the environment corresponding to the traveled distance.

In this paper, we present a Neural Map grOwing system, \textbf{NeMO}, that can digest image sequences, unravel the details of a journey, and produce a comprehensive long-range local map of the environment. Long-range local map is an important input for downstream modules, such as map matching and navigation. In comparison with the \acrshort{bev} maps generated with existing methods, a long-range local map usually has a larger spatial area, requiring fusion of more frames. 

Expanding the spatial size of the \acrshort{bev} plane using current state-of-the-art spatio-temporal fusion methods\cite{li2022bevformer,qin2022uniformer} can be a straightforward approach. However, it is not practical or easily extendable because increased computational costs accompany the expansion of the local map. In \cite{li2022hdmapnet}, an alternative approach is proposed where \acrshort{bev} feature maps for related frames are generated and then aligned onto a fixed \acrshort{bev} space based on ego poses to create a local map, followed by max-pooling for overlapping areas. While this approach is universal and is able to fuse any number of frames, two issues still remain unsettled. First, max-pooling may not eliminate false detections and its performance could suffer if the \acrshort{bev} perception results are poor. The second issue is that accuracy of this approach depends on precise ego poses.

To address these issues, NeMO utilizes a \acrshort{bev}-grid-based local map generation paradigm to fuse frames and construct a long-range local map simultaneously. In NeMO, we create a readable and writable \acrshort{bev} feature grid map for feature storage and extraction, a coarse-to-fine spatial matching module to sample and match features in \acrshort{bev} at different timestamp, along with a homogenous grid fusion network to identify and preserve the most valuable features. The read- and writ-able \acrshort{bev} feature grid map (referred to as ``the big feature map'' in this paper) represents a wider ranged area and stores \acrshort{bev} features, enabling extraction and updating of historical \acrshort{bev} grids whenever required. The system works by initially retrieving \acrshort{bev} features from the current frame. Concurrently, a coarse matching method, is used to sample historical \acrshort{bev} features from the big feature map. To enhance the matching of current and historical features, a finer local spatial matching process is performed. Assuming complete descriptors are present in independent grids, we design a homogeneous grid fusion network to merge the grid-based features. Finally, the corresponding portion of grids in the big feature map is updated with the new fused features. The matching and fusion operations interact in real-time with the big feature map, resulting in an iterative process that enables continuous growth of the neural map. The proposed paradigm offers several advantages. First, it enables the generation of long-range local maps from an arbitrary number of frames at a small and consistent computational cost. Features maintained in the big map integrate all previous information rather than a fixed number of frames. The grid-based coarse-to-fine spatial matching technique mitigates the impact of pose noise, resulting in improved accuracy. The homogenous grid fusion is able to capture, enhance, and update critical information within a grid effectively. It is worth mentioning that NeMO can accommodate a broad spectrum of inputs as it is compatible with any \acrshort{bev} features. A concurrent work\footnote{The first version of the current work was submitted to NeurIPS on 2023-05-17. Shortly after that, we became aware of Xiong et al.\cite{xiong2023neural} CVPR submission which was posted on arxiv on 2023-04-17.} NMP\cite{xiong2023neural} proposes a similar approach like NeMO for better online \acrshort{bev} inference with global map as prior information storage, and we provide a brief comparison in Section~\ref{sec:relatedworks}. 

We validate NeMO system on NuScenes \cite{caesar2020nuscenes} and BDD100K \cite{yu2020bdd100k} datasets. Notably, the latter was acquired using smartphones with reduced accuracy in pose information. To supplement the BDD100K dataset with \acrshort{bev} annotations, we provide annotation tools and use the same annotation style as NuScenes \cite{caesar2020nuscenes} for three categories (lane line, pedestrian crossing, and boundary). The annotated dataset, named BDD-Map, consists of 446 scenes and 426,476 frames.

Our approach demonstrates exceptional performance in the NuScenes dataset, significantly improve performance of HDMapNet \cite{li2022hdmapnet} by a large margin. 
Additionally, our approach shows greater accuracy than the current state-of-the-art temporal \acrshort{bev} perception method BEVerse \cite{zhang2022beverse}, highlighting the flexibility and effectiveness of our system. Furthermore, we provide a benchmark for scene-level local map generation for both datasets.

\begin{figure}
	\includegraphics[width=\linewidth]{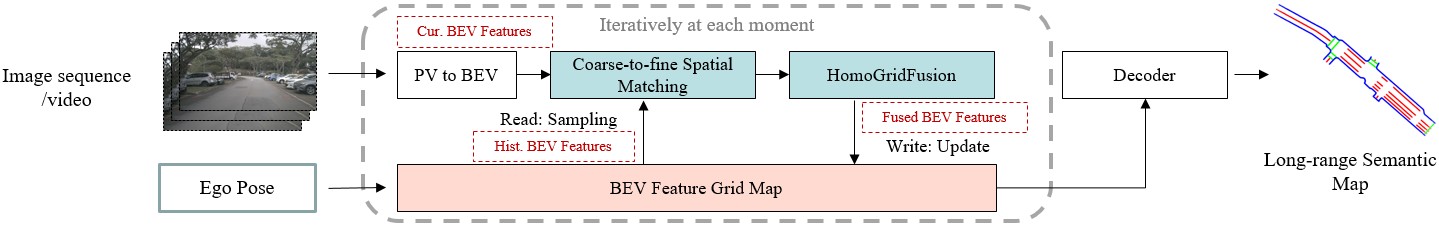}
	\caption{NeMO system overview.}
	\label{fig:nemo-overview}
\end{figure}

\section{Related Work}
\label{sec:relatedworks}
\textbf{\acrshort{bev} lane segmentation map construction.}
The conversion of static road elements in \acrfull{pv} to \acrfull{bev} can be broadly classified into geometry-based and learning-based aproaches. The former utilizes the physical principles underlying the geometric projection relationship between \acrfull{pv} and \acrfull{bev}, while the latter employs data-driven approaches that involve the use of learnable neural networks for mapping. The pioneering geometry-based approach is homography-based IPM (Inverse Perspective Mapping \cite{mallot1991inverse}), which inversely maps \acrshort{pv} information onto \acrshort{bev} plane utilizing homography matrices with flat-ground assumption \cite{garnett20193d,reiher2020sim2real, garnett20193d,mani2020monolayout,can2022understanding}. Therefore, IPM-like methods may have unsatisfactory performance when the ground is not flat. Another geometry-based way, represented by Lift, Splat, and Shoot (LSS) \cite{philion2020lift}, is to lift 2D pixes to 3D space via depth prediction \cite{roddick2018orthographic,gosala2022bird,reading2021categorical,zhang2022beverse}. Learning-based methodologies have made great progress in recent years \cite{lu2019monocular,roddick2020predicting,hendy2020fishing,yang2021projecting,li2022hdmapnet,li2022bevformer,liu2022petrv2,saha2022translating,wang2023bevlanedet}. HDMapNet \cite{li2022hdmapnet} utilizes \acrshort{mlp} to cover complex transformation between \acrshort{pv} and \acrshort{bev} features. Transformers with \acrshort{bev} queries, first used by Tesla \cite{teslaaiday2021} for multi-view \acrshort{pv}-\acrshort{bev} transformation, have gained their popularity in recent works \cite{chen2022persformer,li2022bevformer,liu2022petrv2,peng2023bevsegformer,gong2022gitnet} because of the superior efficacy. In this approach, view transformation is usually conducted using cross-attention between \acrshort{pv} features and \acrshort{bev} queries with positional encoding \cite{li2022bevformer}. As this dense-query design leads to memory cost issue in the cross-attention operation, several studies such as BEVSegFormer \cite{peng2023bevsegformer}, PersFormer \cite{chen2022persformer}, and BEVFormer \cite{li2022bevformer} deploy deformable attention \cite{zhu2020deformable} for faster computation. Concurrently, GKT \cite{chen2022efficient} leverages camera's parameters to find 2D reference points such that queries can focus on small regions. 

\begin{figure}
	\includegraphics[width=\linewidth]{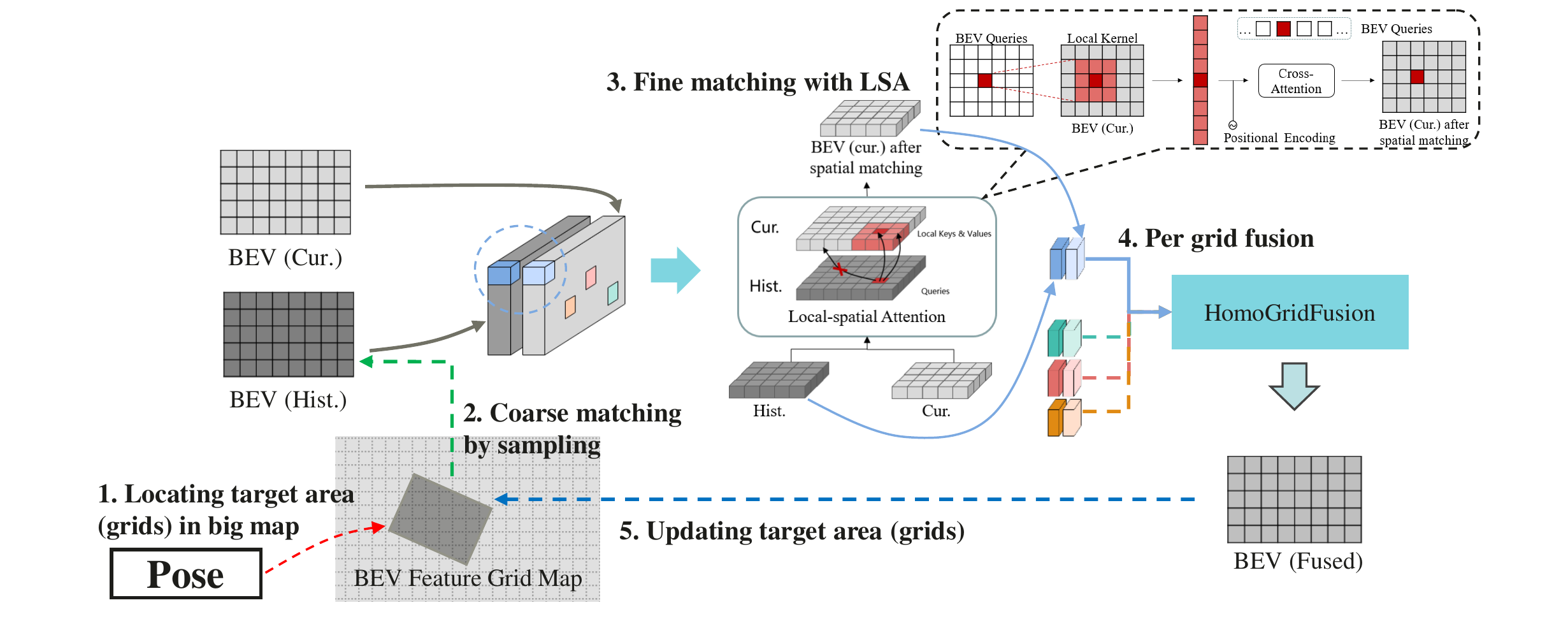}
	\caption{Coarse-to-fine spatial matching and HomoGridFusion models.}
	\label{fig:methodology}
\end{figure}

\textbf{Temporal fusion in \acrshort{bev} lane segmentation.}
Existing studies have confirmed that utilizing multi-frame information helps to improve detection accuracy while alleviating occlusion issue in single frame perception \cite{li2022bevformer,liu2022petrv2,zhang2022beverse,qin2022uniformer}. BEVerse \cite{zhang2022beverse} and BEVFormer \cite{li2022bevformer} wrap past \acrshort{bev} features to the present frame with ego motion information, while the former mainly creates temporal block stack and deploys 3D convolutions, and the latter utilizes self-attention layer to query wrapped previous \acrshort{bev} features with current \acrshort{bev} features. Differently, \cite{liu2022petrv2} and \cite{qin2022uniformer} directly query \acrshort{pv} features. PETRv2 \cite{liu2022petrv2} combines temporal information by adding cooresponding positional encoding for both previous and current image features. Specifically, it generates previous frame's positional encoding by converting its 3D coordinates to the current one according to ego motion, and concatenate the converted 3D coordinates to the 2D features to obtain 3D position-aware features. 3D position-aware features of different frames can be further queried by \acrshort{bev} queries. Unifusion \cite{qin2022uniformer} treats temporal fusion as a multi-view fusion problem by converting past frames to virtual views which are transformed to the ego \acrshort{bev} space with view transformation. The multi-view features are then fused with \acrshort{bev} queries in cross-attention layer. 

Multi-frame fusion is also an inevitable part for long-range local map generation in \acrshort{ads}.
Unifusion \cite{qin2022uniformer} proposes new \acrshort{bev} settings representing larger areas for longer series temporal fusion. However, adopting larger settings has to strike a balance between \acrshort{bev} grid resolution and computational complexity, which may not only impact the result accuracy but also impose constraints on the map size. A scene-level long-term temporal fusion method proposed by HDMapNet \cite{li2022hdmapnet} is to paste \acrshort{bev} maps of previous frames into current's with ego pose and fuse the overlapped grids via max pooling. Yet, fusion methods like temporal max pooling may unexpectedly retain noise thus do not perform stably in different scenes. The real-time long-range local map generation logic proposed by Tesla \cite{teslaaiday2021} is to update only a portion of \acrshort{bev} grids in a long-range map at each moment with ``spatial RNN''. Our work delves further into this logic by proposing a practical pipeline and solution with the aim of aiding future researchers in their exploration of this field.

Most similar to our approach and developed in parallel is NMP \cite{xiong2023neural}, which leverages a city-wide global map for feature prior storage, and a current-to-prior attention followed by ConvGRU module for prior and current feature fusion for online inference. A main difference is that we use a shared recurrent neural network for homogeneous grid fusion, with an assumption that all grid features have identical data distribution pattern regardless of the grid's position in local \acrshort{bev} plane. Besides, we adopt local spatial attention for fine matching, which further reduces the computation cost in online inference process. Moreover, our HDMapNet-based NeMO outperforms \cite{xiong2023neural} in NuScenes dataset for local \acrshort{bev}, and we provide a benchmark for scene-level long-range map generation.

\section{Methodology}
\label{sec:method}

The proposed NeMO system generates a long-range semantic map using image sequence and ego pose information as inputs, as depicted in Figure \ref{fig:nemo-overview}. At each time step, the image $\mathcal{I}_{cur}$ is fed into a \acrshort{pv}-to-\acrshort{bev} neural network, which produces \acrshort{bev} features $\mathcal{F}_{cur}^{bev}$ that are relative to the ego vehicle. The historical \acrshort{bev} features $\mathcal{F}_{hist}^{bev}$ that correspond to the same area are retrieved from the \acrshort{bev} Feature Grid Map (also known as ``the big feature map'') through a ``coarse-to-fine'' spatial matching technique, which we refer to as ``reading''. The \acrshort{bev} features $\mathcal{F}_{cur}^{bev}$ and $\mathcal{F}_{hist}^{bev}$ are then integrated in the HomoGridFusion model to generate $\mathcal{F}_{fused}^{bev}$, and “written” back to the big feature map using the ego pose $E_{cur}$ associated with current moment, which updates the corresponding grids. Finally, the updated big feature map is fed into a decoder to generate the desired local long-range semantic map. 

\textbf{\acrshort{pv}-to-\acrshort{bev} revisiting.} Many studies focus on \acrshort{pv}-to-\acrshort{bev} transformation \cite{li2022hdmapnet,li2022bevformer,reiher2020sim2real,hu2021fiery,zhang2022beverse}, which involves converting images $\mathcal{I}_{cur}$ to \acrshort{bev} features $\mathcal{F}_{cur}^{bev} \subseteq \mathbb{R}_{O_e}^{H_{bev} \times W_{bev} \times K}$ in ego coordinates $O_e$, where $H_{bev}$ and $W_{bev}$ represent the height and width of the \acrshort{bev} plane (i.e., number of grids along height and width), and $K$ is feature dimension. Here the input $\mathcal{I}_{cur}$ can be either one front-view image or multiple images from surrounding cameras. While some research looked into ways to improve ego-centric \acrshort{bev} perception by integrating multiple timestamp frames \cite{li2022bevformer, zhang2022beverse}, ultimately they still produce one single-frame \acrshort{bev} perception map. NeMO is compatible with any \acrshort{pv}-to-\acrshort{bev} frontend as long as it can produce a \acrshort{bev} feature map $\mathcal{F}_{cur}^{bev}$ for current timestamp. 
	
\subsection{Coarse-to-fine spatial matching}
\label{sec:coarsetofine}
	
NeMO system employs a two-stage ``coarse-to-fine'' spatial matching to obtain historical \acrshort{bev} features $\mathcal{F}_{hist}^{bev} \subseteq \mathbb{R}_{O_e}^{H_{bev} \times W_{bev} \times K}$ that correspond to the same spatial area as the $\mathcal{F}_{cur}^{bev}$. 
	
We first use coordinate transformation to obtain a ``coarse'' position for historical \acrshort{bev} features in the readable and writable big feature map $\mathcal{F}^{map}\subseteq \mathbb{R}_{O_g}^{H_{map} \times W_{map} \times K}$, which is in a scene global coordinate system $O_g$. The size of $\mathcal{F}^{map}$, with dimensions $H_{map} \times W_{map}$, is much larger than that of the \acrshort{bev} plane $H_{bev} \times W_{bev}$. The ego pose $E_{cur}$ is critical in identifying the optimal target region within the big feature map for retrieving the historical \acrshort{bev} information. Define grid coordinates in ego-centric \acrshort{bev} plane as $C_{bev}$, their coordinates in the big feature map is coarsely determined as $C_{map} = E_{cur}C_{bev}$. The $C_{map}$ coordinates serve as reference points to sample $\mathcal{F}_{hist}^{bev}$ from $\mathcal{F}^{map}$. 
			
In contrast to established techniques that save previous frames’ features and wrap it to current moment’s coordinate system, potentially leading to loss of information, the coarse spatial matching approach introduced in this study, which employs the big feature map, seamlessly attains spatial alignment for both past and present features with shared scale on the \acrshort{bev} grid plane. Moreover, the big map range ($H_{map}$ and $W_{map}$) is adjustable and can be expanded on demand over time.
			
In ideal situations when $E_{cur}$ is accurate and the grid resolution is high, $\mathcal{F}_{cur}^{bev}$ and $\mathcal{F}_{hist}^{bev}$ are perfectly spatially aligned. However, this is not always the case as sensors may have noises, and grid density needs to be balanced with computational cost. Therefore, we propose the ``fine'' spatial matching stage to alleviate the misalignment issue with a grid-based \acrfull{lsa} network. The \acrshort{lsa} model adopts a local querying approach for each grid by considering only its adjacent grids, rather than querying all grids globally, which not only yields more accurate results but also incurs low computational costs. For a grid $G$ with initialized coarse coordinate $C_{map}^{G}$ in the big feature map, we sample features from $\mathcal{F}_{cur}^{bev}$ using a local kernel that is expanded according to $C_{map}^{G}$, as shown in the upper part of Figure~\ref{fig:methodology}. Denote the sampled features as $\mathcal{F}_{local kernel}^{G}$. \acrshort{bev} queries $Q_{bev}$ are generated by positional encoding in ego-centric \acrshort{bev} plane. Query for $G$, denoted as $Q_{bev}^{G}$ is determined by its position in \acrshort{bev} plane, $C_{bev}^{G}$. For grid $G$, the fine matching current feature is formulated as:
\[\bar{\mathcal{F}}_{cur}^{bev, G} = CA(Q_{bev}^{G}, \mathcal{F}_{localkernel}^{G})\]
where $CA$ refers to cross-attention.
With a query-based structure to integrate information from local regions for each \acrshort{bev} grid, we get $\bar{\mathcal{F}}_{cur}^{bev} \subseteq \mathbb{R}_{O_e}^{H_{bev} \times W_{bev} \times K}$ that better aligned $\mathcal{F}_{hist}^{bev}$.

\subsection{HomoGridFusion}
\label{sec:homogridfusion}
	
$\bar{\mathcal{F}}_{cur}^{bev}$ and $\mathcal{F}_{hist}^{bev}$ is combined with HomoGridFusion, which is a per-grid temporal fusion model of current-historical features. The core spirit of HomoGridFusion is the grid-based shared recurrent network structure. The current and historical \acrshort{bev} features $\bar{\mathcal{F}}_{cur}^{bev}$ and ${\mathcal{F}}_{hist}^{bev} \subseteq \mathbb{R}_{O_e}^{H_{bev} \times W_{bev} \times K}$ represent the same area with the same scale, enabling grid-based temporal fusion at this step.
For each grid $G$ with coordinate $(h, w)$ in \acrshort{bev} plane, where $h \in \{1, 2, ..., H_{bev}\}$ and $w \in \{1, 2, ..., W_{bev}\}$, we can get current \acrshort{bev} feature  $\bar{\mathcal{F}}_{cur}^{bev}[h][w]$ and corresponding historical \acrshort{bev} feature ${\mathcal{F}}_{hist}^{bev}[h][w]$, both of which are $K$-dimentional feature vectors. They are integrated into a new $K$-dimentional feature vectors in a recurrent manner, that ${\mathcal{F}}_{hist}^{bev}[h][w]$ is treated as a hidden state and $\bar{\mathcal{F}}_{cur}^{bev}[h][w]$ is a new observation, and we get the new fused state ${\mathcal{F}}_{fused}^{bev}[h][w]$ with a recurrent model. Since all grids share a same recurrent model, it is easy to parallel in that the \acrshort{bev} feature arrays are unfolded to form a $H_{bev} \times W_{bev}$ sized batch.
This design rests on the assumption that the feature distribution of all \acrshort{bev} grids follows identical pattern, notwithstanding the grid's spatial position $(h, w)$ on the \acrshort{bev} plane $\mathbb{R}_{O_e}^{H_{bev} \times W_{bev} \times K}$. $\mathcal{F}_{cur}^{bev}$ and $\mathcal{F}_{hist}^{bev}$ are embedded in the ego coordinate system which moves as the vehicle advances. $\mathcal{F}_{cur}^{bev}$ represents the features of a specific are based on a single observation, whereas $\mathcal{F}_{hist}^{bev}$ is based on historical observations. The assumption is that the method of combining these two types of features should be the same across different areas regardless of spatial properties.

The main body of the HomoGridFusion model presented in this paper is a two-block bidirectional recurrent network, each followed by an \acrfull{mlp} layer. Prior to the recurrent network blocks, we add three convolutional layers to $\bar{\mathcal{F}}_{cur}^{bev}$ and $\bar{\mathcal{F}}_{hist}^{bev}$ to better capture visual pattern for each grid.

\begin{figure}[]
	\centering
	\includegraphics[width=\linewidth]{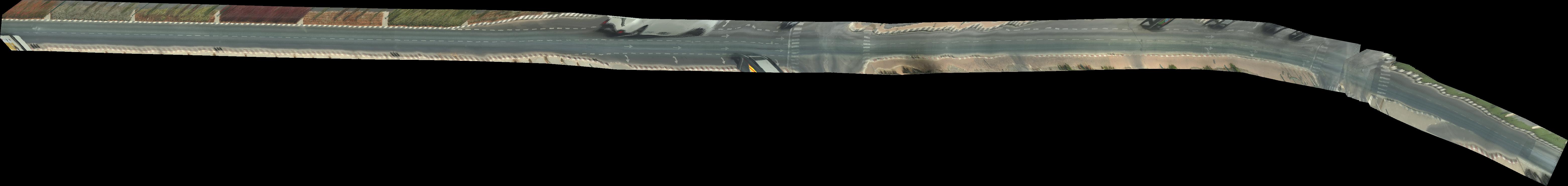}		
	\caption{Pixel map in ground plane.}
	\label{fig:bdd-pixelmap}
\end{figure}

\begin{figure}
	\centering
	\begin{subfigure}[b]{0.3\textwidth}
		\centering
		\includegraphics[width=\linewidth]{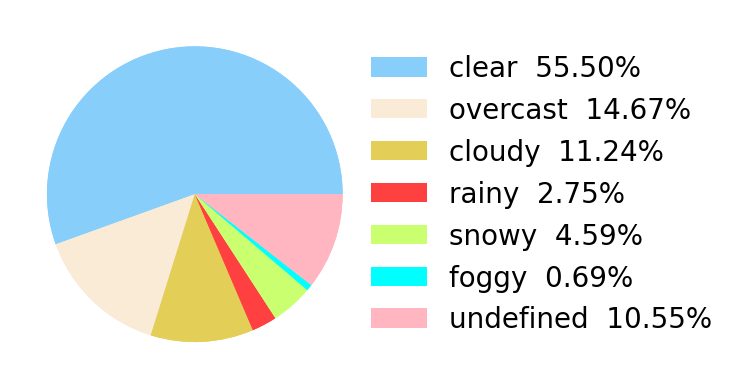}
		\caption{Weather.}
		\label{fig-sub:bdd-weather}
	\end{subfigure}
	\begin{subfigure}[b]{0.3\textwidth}
		\centering
		\includegraphics[width=\linewidth]{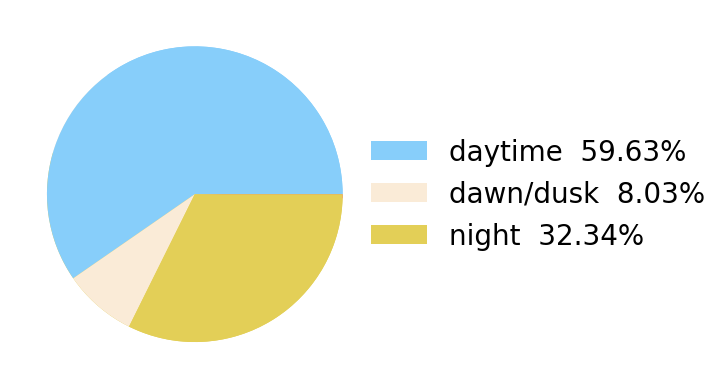}
		\caption{Time of day.}
		\label{fig-sub:bdd-timeofday}
	\end{subfigure}
	\begin{subfigure}[b]{0.3\textwidth}
		\centering
		\includegraphics[width=\linewidth]{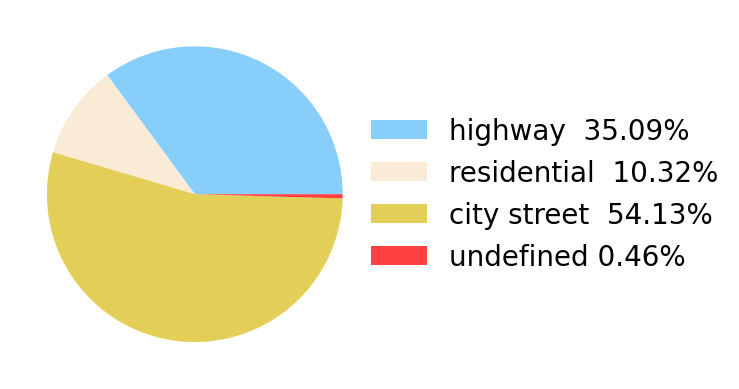}
		\caption{Scene.}
		\label{fig-sub:bdd-scene}
	\end{subfigure}
	\caption{BDD-Map weather condition, time of day, and scene distributions.}
	\label{fig:bdd-dist}
\end{figure}

\section{BDD-Map Dataset}

BDD100K is a large-scale diverse driving video dataset that covers a wide range of driving scenarios, including different times of day and weather conditions in multiple cities. Different from other popular autonomous driving dataset such as NuScenes, Argoverse 2 and Waymo, BDD100K provides less accurate ego motions due to hardware limitation. Instead of using Lidar, BDD100K employs GPS/IMU information from phones to generate rough trajectories. All smartphones are equipped with cameras, GPS and IMU. Therefore, BDD100K represents the most universal perception system for all customers and covers much richer scenes. The computer vision community has greatly benefited from the extensive BDD100K driving scene data. However, its primary usage is for object detection, scene segmentation, and behavior prediction. Few people use it for road structure perception to provide reliable maps. In this work, we attempt to utilize a small portion of it to do semantic map perception task. Although the constructed map accuracy is limited by trajectories precision, they exhibit approximate road topology structure that are sufficient to support lane-level localization and augmented reality road guidance. 

Generating road elements annotations such as lanes, boundaries and pedestrian crossings for a BDD clip in \acrshort{bev} is a challenging task. BDD100K only offers one frame lane annotations per entire clip, leaving out annotations for the majority frames within the clip. Additionally, the dataset does not officially provide extrinsic and intrinsic parameters for each clip, further hindering the process. To tackle these challenges, we develop a semi-automated road elements annotation system. Considering the static nature of road elements and their better observability in a \acrshort{bev}, we propose to convert each 40s video clip into a single large frame as a pixel map as show in figure 1 (a). The pixel map effectively displays all road elements, and obscured areas due to dynamic objects such as cars can be easily annotated with the help of surrounding elements. We provide more details about the semi-automatic annotation pipeline in Appendix A.2.

There are 100 sets in BDD100K and each contains 1000 clips. We randomly selected set 66 and ran the annotation pipeline, resulting in 446 valid ground-plane pixel map with complete road element annotations in \acrshort{bev} and reasonable camera parameters. We omitted the remaining portion due to an unreasonable trajectory or the inability to estimate a suitable extrinsic matrix. We will release the annotated data as an extension of BDD100K, named BDD-Map. The weather condition, scene, and time of day distributions are shown in Figure~\ref{fig:bdd-dist}. Meanwhile, we will release annotation tools since a large portion of BDD100K remains unlabeled and we hope to take advantage of the diversity of BDD100K to promote the development of semantic map perception.

\section{Experiment}

\begin{figure}
	\includegraphics[width=\linewidth]{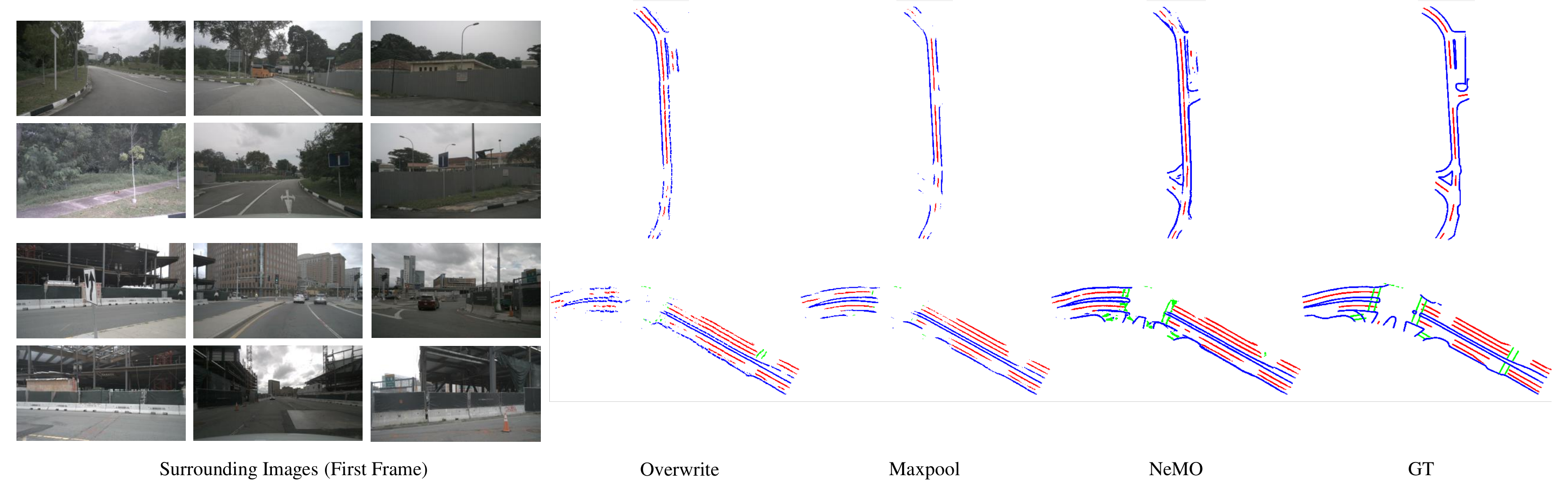}
	\caption{Visual comparison on NuScenes scene-level long-range map generation.}
	\label{fig:result-nus}
\end{figure}

\subsection{Experimental Settings}

\textbf{Datasets and tasks.} We conduct experiments on two open datasets, namely NuScenes (``Nus'' in the following tables) \cite{caesar2020nuscenes} and aforementioned BDD-Map (``BDD'' in the following tables). NuScenes has 1000 scenes. Following HDMapNet \cite{li2022hdmapnet} settings, we use training set (700 scenes) for model training and validation set (150 scenes) for evaluation. BDD-Map has 446 scenes in total, in which 400 scenes are used for training and 46 scenes for evaluation. Regarding the temporal model training, we set frame-per-clip $T=4$ and step-size $D=T=4$ (the distance between the first frames of two consecutive clips when splitting), splitting NuScenes training set into 6923 clips, and BDD-Map dataset into 9565 clips. Following \cite{li2022hdmapnet}, we focus on semantic map segmentation considering road elements of lane line, pedestrian crossing, and boundary. 

\textbf{Experimental settings.} To evaluate NeMO system, we use HDMapNet \cite{li2022hdmapnet} and BEVerse \cite{zhang2022beverse} as the baseline methods. They are served as \acrshort{pv}-to-\acrshort{bev} module in NeMO as well. We conduct experiments under both single front-view image (\textit{\textbf{fcam}}) and six surrounding images (\textit{\textbf{6cam}}) settings for NuScenes dataset, and \textit{\textbf{fcam}} for BDD-Map dataset. 

In \acrshort{pv}-to-\acrshort{bev} process, we uses 30m $\times$ 30m ego plane setting for \textit{\textbf{fcam}} with a 200 $\times$ 200 \acrshort{bev} plane and a resolution of 0.15m, while for \textit{\textbf{6cam}} we adopt the same 30m $\times$ 60m setting (200 $\times$ 400) in \cite{li2022hdmapnet}. In training process, we prepare a fixed-sized plane for each $T$-frame clip in world coordinate system: 256 $\times$ 256 for \textit{\textbf{fcam}} and 384 $\times$ 384 for \textit{\textbf{6cam}} (Table~\ref{tab:single-vs-temp}). 
In inference process, we generate a big map with all frames in each scene, and evaluate the accuracy of these big maps (Table~\ref{tab:temp-fusion}). We use the cross-entropy loss for the semantic segmentation, and Adam optimizer \cite{kingma2014adam} for model training with a learning rate of 1e-03 and weight decay of 1e-07, same with the HDMapNet model \cite{li2022hdmapnet}. We train HDMapNet-based NeMO with one NVIDIA GeForce RTX 3090, and BEVerse-based NeMO with eight. Implementation details in both training and inference phases are presented in Appendix A.1.

\textbf{Evaluation metric and baseline.} For all settings and the big map condition, \acrfull{miou} is used as the evaluation metric. HDMapNet \cite{li2022hdmapnet} and BEVerse\cite{zhang2022beverse} are selected as the baselines for per timestamp \acrshort{bev}  perception comparison. A major approach of generating big maps from multiple \acrshort{bev}s is stitching single-frame \acrshort{bev}s with ego poses and integrate the overlapped grids. We select two common non-parameterized methods, overwriting the grids by the information captured in the most recent moment or keeping the maximum (maxpool in temporal dimension), as baselines to compare with our neural network temporal fusion way.

\begin{table}[]
	\caption{Experiments of \acrshort{bev} segmentation without and with temporal fusion, for both single front view (\textit{\textbf{fcam}}) and six surrounding image (\textit{\textbf{6cam}}) inputs. Three segmentation classes are road divider (Divider), pedestrian crossing (Ped Xing), and boundary (Boundary). $*$ means the results are reported in the paper \cite{li2022hdmapnet, zhang2022beverse}. $\dag$ means the results are reimplemented in this work.}
	\label{tab:single-vs-temp}
	\vspace{1em}
	\centering
	\begin{tabular}{@{}llcccccc@{}}
		\toprule
		\textit{\textbf{fcam}} &                    &  &   & \multicolumn{4}{c}{\acrshort{miou}(\%)} \\                 
		&                    & \#Param. &  Temp    & Divider & \multicolumn{1}{l}{Ped Xing} & \multicolumn{1}{l}{Boundary} & \multicolumn{1}{l}{All} \\ \cmidrule(l){5-8} 
		\multirow{3}{*}{Nus} 
		& HDMapNet$_{200 \times 200}$$\dag$ &  27.0M &  & 38.4      & 18.7        & 35.3       & 30.8  \\
		& NeMO$_{200 \times 200}$     &  &  \checkmark & 40.9$_{\textcolor{blue}{+2.5}}$     & 23.1$_{\textcolor{blue}{+4.4}}$                              & 39.4$_{\textcolor{blue}{+4.1}}$                           & 34.5$_{\textcolor{blue}{+3.7}}$                     \\
		& NeMO$_{256 \times 256}$    &  28.2M & \checkmark  & 42.8      & 21.2                               & 40.8                           & 34.9                      \\ \cmidrule(l){2-8} 
		\multirow{3}{*}{BDD} 
		& HDMapNet$_{200 \times 200}$$\dag$ &  23.8M &    &  24.3      & 7.2           & 14.3       & 15.3  \\
		& NeMO$_{200 \times 200}$     &  & \checkmark  & 26.7$_{\textcolor{blue}{+2.4}}$      & 10.1$_{\textcolor{blue}{+2.9}}$           & 17.2$_{\textcolor{blue}{+2.9}}$       & 18.0$_{\textcolor{blue}{+2.7}}$  \\ 
		& NeMO$_{256 \times 256}$    & 25.0M  & \checkmark   & 28.4      & 8.3                               & 15.4                           & 17.4                      \\ \cmidrule(l){1-8} 
		\textit{\textbf{6cam}}         &           &                    &      &         &                                  &                              &                         \\
		\multirow{6}{*}{Nus} 
		& HDMapNet$_{200 \times 400}$$*$ & 78.3M &      & 40.6      & 18.7        & 39.5       & 32.9  \\
		& NeMO$_{200 \times 400}$    &   & \checkmark  & 45.9$_{\textcolor{blue}{+5.3}}$      & 26.9$_{\textcolor{blue}{+8.2}}$           & 46.0$_{\textcolor{blue}{+6.5}}$       & 39.6$_{\textcolor{blue}{+6.7}}$  \\  
		& NeMO$_{384 \times 384}$   &  79.5M  & \checkmark  & 44.7      & 22.9                               & 44.5                           & 37.4                       \\ \cmidrule(l){2-8} 
		& BEVerse-Map$_{200 \times 400}$$*$  & 54.8M  &      & 53.9      & 41.0                               & 54.5                           & 49.8                      \\
		& NeMO$_{200 \times 400}$   &    & \checkmark  & 57.7$_{\textcolor{blue}{+3.8}}$      & 47.8$_{\textcolor{blue}{+6.8}}$                               & 57.6$_{\textcolor{blue}{+3.1}}$                           & 54.4$_{\textcolor{blue}{+4.6}}$                      \\
		& NeMO$_{384 \times 384}$   &  55.5M   & \checkmark  & 57.1      & 47.0                               & 57.6                           & 53.9                      \\ \bottomrule
	\end{tabular}
\end{table}

\subsection{Main Results}
\label{sec:results}

\begin{table}[]
	\caption{Experiments of \acrshort{bev} stitching and temporal fusion methods in multi-frame grid fusion to generate long-range maps, for both single front view (\textit{\textbf{fcam}}) and six surrounding image (\textit{\textbf{6cam}}).}
	\label{tab:temp-fusion}
	\vspace{1em}
	\centering
	\begin{tabular}{@{}lllcccc@{}}
		\toprule
		& \multirow{2}{*}{\begin{tabular}[c]{@{}l@{}}Single-frame \\ \acrshort{pv}-to-\acrshort{bev}\end{tabular}} & \multirow{2}{*}{\begin{tabular}[c]{@{}l@{}}Multi-frame \\ Grid Fusion\end{tabular}} & \multicolumn{4}{c}{\acrshort{miou}(\%)}             \\
		\textit{\textbf{fcam}}                         &                                                                                    &                                                                                 & Divider & Ped Crossing & Boundary & All \\ \cmidrule(l){4-7} 
		\multirow{3}{*}{Nus}    & \multirow{3}{*}{HDMapNet}                                                          & Overwrite                                                                       & 40.7      & 13.5           & 36.4       & 30.2  \\
		&                                                                                    & Maxpool                                                                             & 42.6      & 12.7           & 39.4       & 31.5  \\
		&                                                                                    & NeMO                                                                             & \textbf{47.2}      & \textbf{21.6}           & \textbf{44.8}       & \textbf{37.9}  \\ \cmidrule(l){3-7} 
		\multirow{3}{*}{BDD} & \multirow{3}{*}{HDMapNet}                                                          & Overwrite                                                             & 29.9      & 7.0           & 16.4       & 17.8  \\
		&                                                                   & Maxpool                                                                   & 26.2      & 4.7           & 14.9       & 13.9  \\
		&                                                                   & NeMO                                                                   & \textbf{37.5}      & \textbf{12.8}           & \textbf{22.5}       & \textbf{24.3}  \\ \midrule
		\textit{\textbf{6cam}}                         &                                                                                    &                                                                                 &         &              &          &     \\
		\multirow{6}{*}{Nus}    & \multirow{3}{*}{HDMapNet}                                                          & Overwrite                                                             & 34.4      & 10.0           & 32.1       & 25.6  \\
		&                                                                                    & Maxpool                                                                   & 43.3      & 13.9           & 42.1       & 33.1  \\
		&                                                                                    & NeMO                                                                   & \textbf{48.7}      & \textbf{29.1}           & \textbf{49.7}       & \textbf{42.5}  \\ \cmidrule(l){3-7} 
		& \multirow{3}{*}{BEVerse}                                                           & Overwrite                                                                       & 52.6      & 28.1           & 49.3       & 43.3  \\
		&                                                                                    & Maxpool                                                                             & 61.7      & 46.1           & 59.8       & 55.8  \\
		&                                                                                    & NeMO                                                                             & \textbf{62.5}      & \textbf{49.5}           & \textbf{61.6}       & \textbf{57.9 } \\ \bottomrule
	\end{tabular}
\end{table}

\begin{table}[]
	\caption{Experiments of NeMO (a) w/ and w/o \acrshort{lsa} fine spatial matching, (b) different supervision types, and (c) differen designs of HomoGridFusion. Results shown in the table are \acrshort{miou} (\%). For comparison, the first column shows the results of NeMO system with \acrshort{lsa} model, many-to-one supervision, and has 2d convolutional layers in HomoGridFusion.}
	\label{tab:ablation_experiments}
	\vspace{1em}
	\centering
	\begin{tabular}{@{}lcccccc@{}}
		\toprule
		&  & & Spatial Matching & Supervision  & \multicolumn{2}{c}{HomoGridFusion Design} \\
		& Baseline & NeMO     & w/o \acrshort{lsa}           & Many-to-many & LSTM             & Conv1d+LSTM            \\ \cmidrule(l){4-7} 
		Divider & 43.3 & 48.7     & 46.8             & 42.0         & 44.7             & 45.1                   \\
		Ped Xing & 13.9 & 29.1     & 27.9             & 20.0         & 20.7              & 22.0                    \\
		Boundary & 42.1 & 49.7     & 46.2             & 42.5         & 44.8             & 43.9                   \\
		All      & 33.1 & 42.5     & 40.3             & 34.9         & 36.8             & 37.0                   \\ \bottomrule
	\end{tabular}
\end{table}

Table~\ref{tab:single-vs-temp} presents results of \acrshort{bev} segmentation performance before and after fusion with NeMO. To provide ego-centric \acrshort{bev} perception on $200\times200$ or $200\times400$ settings evaluation, features in NeMO are extracted back from the big feature map to the ego-centric plane according to ego pose at each moment. For \textit{\textbf{fcam}}, multi-frame fused NeMO outperforms the baseline, single-frame HDMapNet\cite{li2022hdmapnet}, in both NuScenes and BDD-Map datasets. For NuScenes \textit{\textbf{6cam}}, NeMO improve HDMapNet \acrshort{miou} by 20.36\%, from 32.9 to 39.6. This suggests that the proposed strategy NeMO successfully enhances perception in map growing process by integrating temporal information in consecutive frames. For a more advance baseline BEVerse \cite{zhang2022beverse}, NeMO also improves the per timestamp perception \acrshort{miou} from 49.8 to 54.4. 
It is worth mentioning that NeMO only introduce a minor increase in model size, meaning that NeMO has potential of obtaining considerable advantages at an inconsequential cost.\footnote{The reimplementation of \textit{\textbf{fcam}} HDMapNet exhibits a significantly lower parameter count than its official counterpart \textit{\textbf{6cam}} HDMapNet. This discrepancy can be attributed to the difference in view fusion module architecture between the two models, wherein the latter employs 6 independent \acrshort{mlp}s for 6 surrounding camera views while the former only requires one \acrshort{mlp} for \textit{\textbf{fcam}}.}. 

We also generate big local map for each scene (150 scenes in NuScenes and 46 scenes in BDD-Map) and evaluate the scene-based map segmentation performance in Table~\ref{tab:temp-fusion}. The results demonstrate that NeMO leads to significantly improved accuracy of perception when compared to non-parameterized baseline techniques, across all conditions. This suggests that the proposed design of the grid-based fusion architecture and shared fusion model can effectively handle the processes of information screening, updating, and memorization involved in temporal fusion. Besides, it can be observed Maxpool outperform Overwrite in all settings except for the case of BDD \textit{\textbf{fcam}}. It is due to the low accuracy of the pose in BDD-Map dataset: selecting the maximum value tends to maintain more noise data which in turn diminishes the overall accuracy.

\subsection{Ablation Study}
\label{sec:ablation-study}

We conduct ablation study based on HDMapNet-NeMO using \textit{\textbf{6cam}} NuScenes dataset. 

\textbf{Impacts of \acrfull{lsa} model.}
We analyze the impact of fine spatial matching \acrshort{lsa} model in NeMO system. As shown in Table~\ref{tab:ablation_experiments}, NeMO with \acrshort{lsa} model demonstrates advantages in all classes compared to the result without \acrshort{lsa}. This experiment highlights the role of local-spatial fusion design in enhancing the performance of temporal fusion in map generation.

\textbf{Convolutional layers in HomoGridFusion.} For the default setting, we use 2D convolutional layers before \acrshort{lstm}\cite{hochreiter1997long} recurrent block. We compare it with two other designs in HomoGridFusion, i.e., HomoGridFusion without convolutional layers (``\acrshort{lstm}'') and with simple Conv1D (``conv1d+\acrshort{lstm}''). Table~\ref{tab:ablation_experiments} shows that 2D convolutional layers lead to better performance for long-range map generation.

\textbf{Supervision types in HomoGridFusion.} The recurrent network in HomoGridFusion can be supervised in various ways. Many-to-many supervision is implemented by conducting supervision to partial grids at each timestamp, while many-to-one supervision applies clip-based map-wide supervision to all grids evolved. Results in Table ~\ref{tab:ablation_experiments} demonstrate that many-to-one supervision surpasses the performance of many-to-many method, suggesting that many-to-one supervision enables the model to learn to effectively disregard and memorize information across multiple frames in a more global sense. Such an approach is particularly beneficial for NeMO's map generation process.

\begin{table}[]
	\caption{Experiments of map generation with noisy pose information.}
	\label{tab:pose_noises}
	\centering
	\small
	\vspace{1em}
	\begin{tabularx}{\textwidth}{l*{9}{C}}
		\hline
		Noise    & \multicolumn{3}{c}{{[}0, 0{]}} & \multicolumn{3}{c}{{[}0.1, 0.1{]}} & \multicolumn{3}{c}{{[}0.5, 0.5{]}} \\
		& Overwrite   & Maxpool  & NeMO  & Overwrite    & Maxpool    & NeMO   & Overwrite    & Maxpool    & NeMO   \\ \cline{2-10} 
		Divider  & 34.4        & 43.3     & 48.7  & 33.2         & 42.2       & 48.0   & 22.5         & 15.8       & 39.6   \\
		Ped Xing & 10.0        & 13.9     & 29.1  & 9.9          & 13.6       & 28.8   & 8.2          & 8.5        & 26.8   \\
		Boundary & 32.1        & 42.1     & 49.7  & 31.2         & 41.5       & 49.1   & 23.1         & 23.9       & 44.1   \\
		All      & 25.5        & 33.1     & 42.5  & 24.8         & 32.4       & 42.0   & 17.9         & 16.1       & 36.8   \\ \hline
	\end{tabularx}
\end{table}

\textbf{Impacts of pose noises.} We validate NeMO’s capability in handling noisy pose information through experiments on the BDD-Map dataset in Table~\ref{tab:single-vs-temp} and Table~\ref{tab:temp-fusion}. Besides, we manually introduce pose noises in the NuScenes dataset and compare NeMO with baselines across two noise levels. In Table~\ref{tab:pose_noises}, we show results of Gaussian noise added to ego pose with different standard deviation. Specifically, $[0.5, 0.5]$ represents a random $e \in N(0, 0.5)$ degree is added to each of three Euler angles in R, and $e \in N(0, 0.5)$ meters of noise are added to the x and y coordinates in T. It can be observed that, as the pose noise increases, the performance of long-range maps generated via overwriting and maxpooling substantially deteriorates. Despite a slight reduction, NeMO exhibits a relatively high level of performance. Notably, even with a 0.5 noise level present, NeMO outperforms the noise-free maxpool method, demonstrating its considerable capability to tackle noisy poses.

\section{Conclusion and Discussion}
\label{sec:conclusion}

\textbf{Conclusion.} This paper presents a novel neural map growing system, named NeMO. By employing coarse-to-fine spatial matching and HomoGridFusion module to fuse temporal information, NeMO generates long-range segmentation maps of target areas from image streams. Experiments demonstrate that, with model size increase a little for temporal fusion, NeMO leads to a significant improvement in the generated \acrfull{bev} map compared to other existing methods. Besides, we evaluate the performance of long-range local maps generated from all images in each scene and provide a benchmark within this novel evaluation framework. We show that NeMO achieves broad generalization across scenes and various sizes of \acrshort{bev} plane. Meanwhile, we extend a portion of BDD100K dataset by incorporating \acrshort{bev} map element annotations and release BDD-Map as a new \acrshort{bev} dataset. We hope to provide a comprehensive and diverse resource to facilitate further advancement in the field of related research.

\textbf{Broader impacts.}The study suggests that employing a shared, grid-based, and location- and view-independent fusion network to temporally fuse individual \acrshort{bev} grids in a big feature map yields significant improvements. Instead of fusing temporal \acrshort{bev} maps in ego coordinate system, it extracts, denoises, memorizes, and updates map information in real space, which redefine the paradigm of temporal fusion. We release a BDD-Map dataset and tools for generating \acrshort{bev} annotations to aid others in producing more diverse \acrshort{bev} datasets. We hope these initial exploratory undertakings and related resources will advance \acrshort{bev} perception. 

\textbf{Limitations and future directions.} One of the limitations of this study is evident in the implementation aspect. While NeMO framework supports end-to-end training, a two-stage approach is employed in this study, whereby the single-frame \acrshort{pv}-to-\acrshort{bev} perception model and the multi-frame fusion model are separately trained and supervised. As a result, the ultimate fusion outcome is notably constrained by the initial \acrshort{pv}-to-\acrshort{bev} model, limiting its performance and effectiveness. Therefore, a potential avenue for future research is to investigate methodologies for end-to-end training.

\bibliographystyle{plain}
\bibliography{ref}

\appendix

\section{Appendix}

\subsection{Implementation Details}
\label{appndx:implementation}

The workflow of the NeMO system incorporates three stages: \acrshort{pv}-to-\acrshort{bev} feature transformation, coarse-to-fine spatial matching, and temporal fusion with HomoGridFusion. Corresponding \acrshort{pv}-to-\acrshort{bev} and spatial matching \& temporal fusion sub-networks may be subject to end-to-end training or separate training. This paper follows the latter scheme and present details regarding the implementation of training and inference. 
The system and models we employ in this study are designed for semantic segmentation task, aiming to recognize three static map elements: lane divider, pedestrian crossing, and boundary.

\subsubsection{Training phase} 

The data loaders used for \acrshort{pv}-to-\acrshort{bev} model and spatial matching \& temporal fusion model (\acrshort{lsa} and HomoGridFusion) are different as the latter requires time-series image sequence rather than single frame image. In the case of \acrshort{pv}-to-\acrshort{bev} model, each sample consists of images obtained at one timestamp. The \acrshort{pv}-to-\acrshort{bev} model itself is trained in an end-to-end manner. We omit relevant descriptions as we conduct the same operations as elabrated in corresponding studies, e.g., HDMapNet\cite{li2022hdmapnet} and BEVFormer\cite{zhang2022beverse} in the current work. In the following, we focus on HDMapNet-based NeMO with \textit{\textbf{6cam}} input and provide more implementation details in practice.

\textbf{Clip data preparation.} For \acrshort{lsa} and HomoGridFusion network, a sample from data loader is composed of a clip created with a sequence of $T=4$ consecutive frames, and a batch has $B$ 4-frame clips in total. Batch size $B\in \{1, 2, 4, 8\}$ is adjustable in distributed training. For instance, we use $B=8$ when training with 8 GPUs. 
In each clip, each image $\mathcal{I}_t$ is first converted to $\mathcal{F}_{t}^{bev} \subseteq \mathbb{R}_{O_e}^{200 \times 400 \times 4}$, $t \in \{1, 2, ..., T\}$, via the trained \acrshort{pv}-to-\acrshort{bev} model HDMapNet. Here HDMapNet is in evaluation mode and the parameters are not updated in the process. While $\mathcal{F}_{t}^{bev}$ can be \acrshort{bev} features obtained at any stage in the \acrshort{pv}-to-\acrshort{bev} network, we use the features returned by semantic HDMapNet decoder as $\mathcal{F}^{bev}_t \subseteq \mathbb{R}_{O_e}^{200 \times 400 \times 4}$ in ego-centric coordinate $O_e$. 

\begin{figure}
	\includegraphics[width=\linewidth]{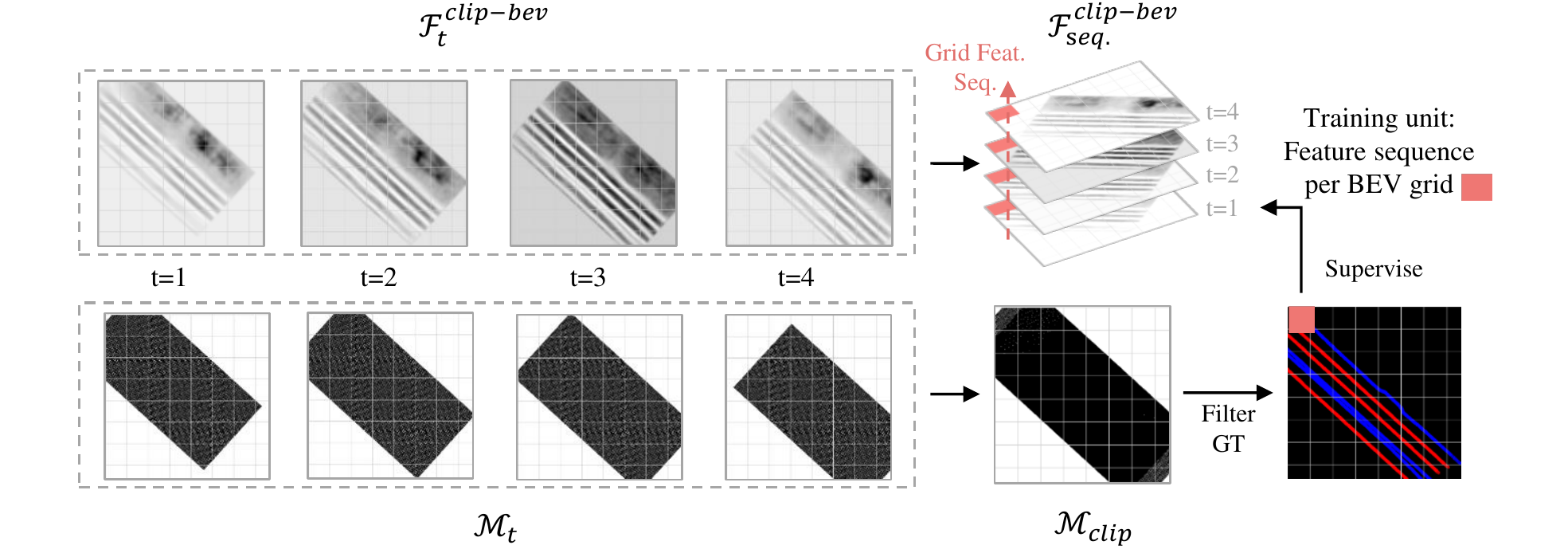}
	\caption{Temporal fusion model training sample generation process.}
	\label{fig:training}
\end{figure}

For each clip, we restrict the spatial range of interest in $O_g$ by defining a big \acrshort{bev} plane $\mathcal{F}_{clip}^{bev} \subseteq \mathbb{R}_{O_g}^{384 \times 384 \times 4}$ functioning as the big \acrshort{bev} feature map. At each timestamp $t \in \{1, 2, 3, 4\}$, ego-centric $\mathcal{F}_{t}^{bev} \subseteq \mathbb{R}_{O_e}^{200 \times 400 \times 4}$ can be mapped to generate $\mathcal{F}_{t}^{clip-bev} \subseteq \mathbb{R}_{O_g}^{384 \times 384 \times 4}$ with ego pose $E_t$ (see four $\mathcal{F}_{t}^{clip-bev}$ figures in Figure~\ref{fig:training}). Meanwhile, we generate masks $\mathcal{M}_{t}$ corresponding to the mapped location in the big \acrshort{bev} plane for loss calculation.

\textbf{Fine spatial matching.} During NeMO experimental trials conducted without \acrshort{lsa} architecture (Section 5.3 and Table 3), the sequence $\mathcal{F}_{T}^{clip-bev}$ is directly fed to temporal fusion (HomoGridFusion) model as shown in Figure~\ref{fig:training}. When incorporating fine spatial matching, we add a \acrlong{lsa} stucture as we described in Section 3.1 at this stage. For timestamp $t\in \{2, 3, 4\}$, the better aligned features $\bar{\mathcal{F}}^{bev}_{t, cur}$ returned from \acrshort{lsa}, instead of the original $\mathcal{F}^{bev}_{t, cur}$, are converted to big \acrshort{bev} plane with ego pose $E_t$ to obtain $\mathcal{F}_{t}^{clip-bev}$. 

Besides, Gaussian blurring or linear interpolation can be added to $\mathcal{F}_{t}^{bev}$ for smoothing. In practice, we use Gaussian burring (with kernel size 7 and sigma 5) considering the processing time.

\textbf{Temporal fusion with HomoGridFusion.} 
As shown in Figure~\ref{fig:training}, all four $\mathcal{F}_{t}^{clip-bev}$ in each clip can form a temporal sequence of \acrshort{bev} features $\mathcal{F}_{seq}^{clip-bev} \subseteq \mathbb{R}_{O_g}^{B \times 4 \times 384 \times 384 \times 4}$. 
In one clip, each grid in the big \acrshort{bev} plane, denoted by $\mathcal{F}_{seq}^{clip-bev}[h][w]$, is a sequence of $4$ historical \acrshort{bev} features. For HomoGridFusion tructure, grid feature sequence is the sample unit for training, and a $\mathcal{F}_{seq}^{clip-bev}$ has $384 \times 384$ sample units. We illustrate one grid feature sequence in Figure~\ref{fig:training} with pink squares. 
The ground truth $\mathcal{F}_{GT} \subseteq \mathbb{R}^{B \times 384 \times 384 \times 4}$ is first cropped according to the global location and area of $\mathcal{F}_{seq}^{clip-bev}$, and then filtered with the assistance of mask $\mathcal{M}_{clip}$ generated by combining all $\mathcal{M}_{t}$s. In each clip, the corresponding ground truth grid $\mathcal{F}_{GT}[h][w]$ is used to supervise sample unit $\mathcal{F}_{seq}^{clip-bev}[h][w]$ in HomoGridFusion. 

In practice, we first add three $3 \times 3$ convolutional layers to $\mathcal{F}_{seq}^{clip-bev}$ with strides of $1$ and output channels of $64, 128, 256$. The first two convolutional layers are followed by ReLU and $3\times 3$ maxpooling layers with strides of $1$ and paddings of $1$, and the last convolutional layer followed by batch normalization and ReLU. The outcome of the convolutional layers $\mathcal{F}_{seq}^{clip-bev}$ has a shape of $B \times 4 \times 384 \times 384 \times 256$, where $B$ is batch size, $4$ is sequence length, and $256$ is feature channels. We adopt two bidirectional \acrshort{lstm} followed by two \acrshort{mlp} structures for grid-based feature sequence fusion in HomoGridFusion. For \acrshort{lstm} training, the $\mathcal{F}_{seq}^{clip-bev}$ is unfolded to generate $B \times 384 \times 384$ feature sequences. The input batch size of the model was configured to be $B \times 384 \times 384$, while the sequence length and number of feature channels are set to $4$ and $256$, respectively. In \acrshort{lstm} temporal fusion, we reduce the resultant output feature channels back to 4.
With many-to-one strategy, we use $\mathcal{F}_{GT}$ to supervise the grid features in the last sequence layer of \acrshort{lstm} outcomes.

\subsubsection{Inference phase}

The inference process is similar to the procedure shown in Figure 1. In inference phase, we first define an all-zero initial $\mathcal{F}_{map}^{bev} \subseteq \mathbb{R}_{O_g}^{H_{bev}^{map} \times W_{bev}^{map} \times K}$ representing the big feature map. 

We process each frame in a scene in chronological order. For each frame at timestamp $t$, $\mathcal{I}_t$ is converted to $\mathcal{F}_{cur}^{bev} \subseteq \mathbb{R}_{O_e}^{200 \times 400 \times 4}$. 
We conduct coarse spatial matching to obtain historical \acrshort{bev} $\mathcal{F}_{hist}^{bev} \subseteq \mathbb{R}_{O_e}^{200 \times 400 \times 4}$ by sampling from $\mathcal{F}_{map}^{bev}$ according to ego pose $E_t$, and fine spatial matching to obtain $\bar{\mathcal{F}}_{cur}^{bev} \subseteq \mathbb{R}_{O_e}^{200 \times 400 \times 4}$ with \acrshort{lsa} model. $\bar{\mathcal{F}}_{cur}^{bev}$ and $\mathcal{F}_{hist}^{bev}$ are temporally fused with HomoGridFusion to obtain $\mathcal{F}_{fused}^{bev}$, which is further used to update corresponding grids in the big feature map $\mathcal{F}_{map}^{bev}$. 

Finally, for evaluation, we use one hot encoding to convert $\mathcal{F}_{map}^{bev}$ to long-range local map.


\subsection{BDD-Map Annotation Pipeline}
\label{appndx:bdd-pipeline}

\begin{figure}
	\centering
	\begin{subfigure}[b]{0.25\textwidth}
		\centering
		\includegraphics[width=\linewidth]{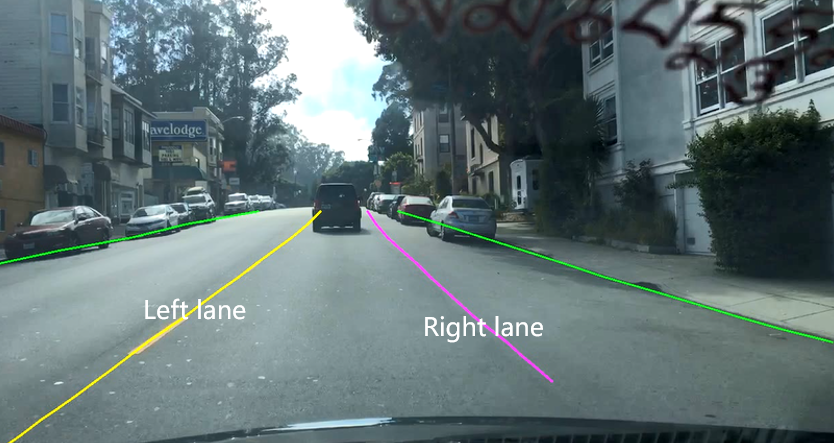}
		\caption{Key frame example.}
		\label{fig-sub:bdd-keyframe}
	\end{subfigure}
	\begin{subfigure}[b]{0.65\textwidth}
		\centering
		\includegraphics[width=\linewidth]{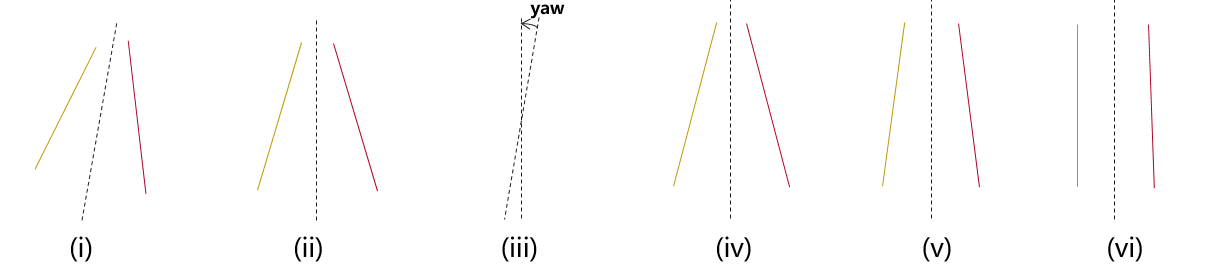}
		\caption{Yaw and pitch angle adjustment illustration.}
		\label{fig-sub:bdd-yawpitch}
	\end{subfigure}
	\caption{BDD-Map clip-level extrinsic adjustment.}
	\label{fig:bdd-extrinsic-adjustment}
\end{figure}

\begin{figure}[]
	\centering
	\begin{subfigure}[b]{\textwidth}
		\centering
		\includegraphics[width=\linewidth]{pic/bdd1a.jpg}
		\caption{Pixel map in ground plane.}
		\label{fig:bdd-pixelmap}
	\end{subfigure}
	
	\begin{subfigure}[b]{\textwidth}
		\centering
		\includegraphics[width=\linewidth]{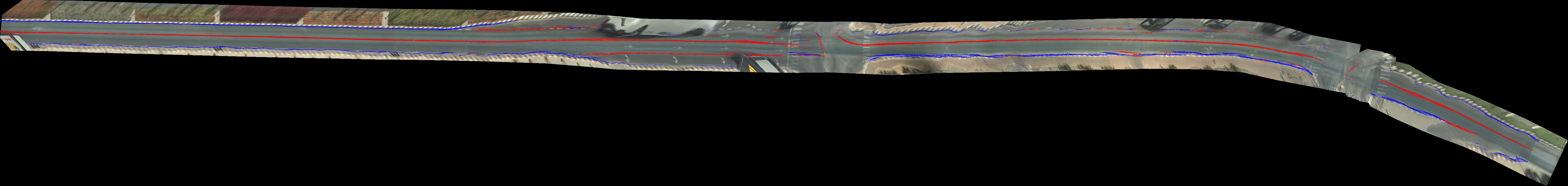}
		\caption{Pixel map with projected annotations by a trained model in ground plane. Blue points represent boundaries and red points represent lanes.}
		\label{fig:bdd-projectedannotation}
	\end{subfigure}
	
	\begin{subfigure}[b]{\textwidth}
		\centering
		\includegraphics[width=\linewidth]{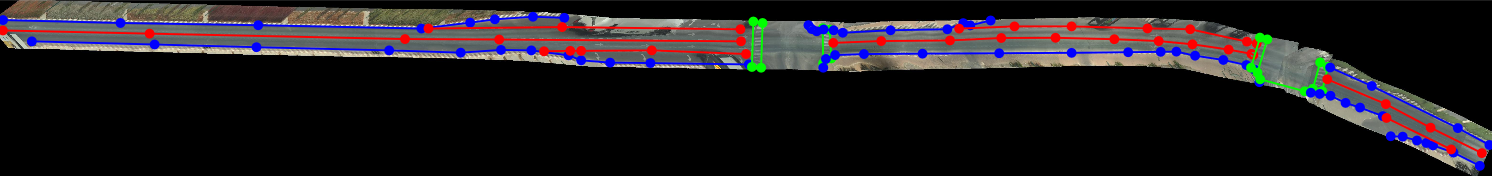}
		\caption{Pixel map with projected annotations after manual adjustment in ground plane. Blue lines represent boundaries, red lines represent lanes, and green lines represent pedestrian crossings.}
		\label{fig:bdd-mannuallyadjustment}
	\end{subfigure}
	\caption{Semi-automated \acrshort{bev} annotation generation of one clip.}
	\label{fig:bdd-semiautomated}
\end{figure}

For each 40s clip, BDD100K provides a rough trajectory in 1 Hz. It is interpolated into 30Hz and subsequently every frame in the clip video is assigned a rough pose based on the trajectory. Videos in BDD100K are captured by iPhone 5 with 720p videos, and therefore we approximate intrinsic parameters based on focal length 700px and assume no distortion. To address the variability of unrestricted phone placements by drivers, we propose a pipeline for estimating extrinsic parameters, which may differ from clip to clip.

Based on the observation that phone is not likely repositioned significantly within one clip, we use a constant extrinsic matrix for each clip. To begin, we utilize a well-trained \acrshort{pv} lane detection model to extract the lanes and boundaries within each frame of the video. From there, we select key frames that have two clearly detected lane lines on both sides of the ego car, left lane and right lane as shown in Figure~\ref{fig-sub:bdd-keyframe}. Assuming that the lane lines on both sides of the lane in which the car is traveling are parallel, we adjust the pitch and yaw angles in the extrinsic parameters until the lane lines on both sides of the vehicle in \acrshort{bev} are parallel, and we assume a roll angle of zero for all clips based on the observed vertical stability of the scene in the videos. We first initialize pitch and yaw as zeros, and cast the left and right lane into \acrshort{bev}. Due to yaw angle, two parallel lanes in \acrshort{bev} are not symmetric about a vertical axis as shown in Figure~\ref{fig-sub:bdd-yawpitch}(i). 
We calculate the axis of symmetry for the two lanes in \acrshort{bev}, and rotate it to align with the vertical axis (Figure~\ref{fig-sub:bdd-yawpitch}(ii)). The resulting angle of rotation is referred to as the yaw angle (Figure~\ref{fig-sub:bdd-yawpitch}(iii)).
We then adjust the pitch angle iteratively to make the symmetric lanes parallel to the vertical axis as shown in Figure~\ref{fig-sub:bdd-yawpitch}(v) and Figure~\ref{fig-sub:bdd-yawpitch}(vi). 

With ego poses per frame, constant intrinsic matrix and optimized extrinsic matrix, we construct the ground plane pixel map with Inverse Perspective Mapping (IPM) algorithm. For overlapping area, pixels are overwritten by the most recent frame. Meanwhile, road elements extracted by a model are transformed to the ground plane as shown in Figure~\ref{fig:bdd-projectedannotation}. The road elements related points are filtered and vectorized into instance level lanes and boundaries as shown in Figure~\ref{fig:bdd-mannuallyadjustment}.

To ensure accuracy, the final annotated results are subjected to human inspection and adjustments, and pedestrian crossings are manually added to the annotations. In order to facilitate human inspection and provide additional information that may not be clear in the \acrshort{bev}, we have developed a special command based on LabelMe\cite{russell2005labelme}. When a user right-clicks on a point in the pixel map, the corresponding image will appear for further inspection.

\subsection{Visualization of Experimental Results in NuScenes and BDD-Map}

\begin{figure}
	\includegraphics[width=\linewidth]{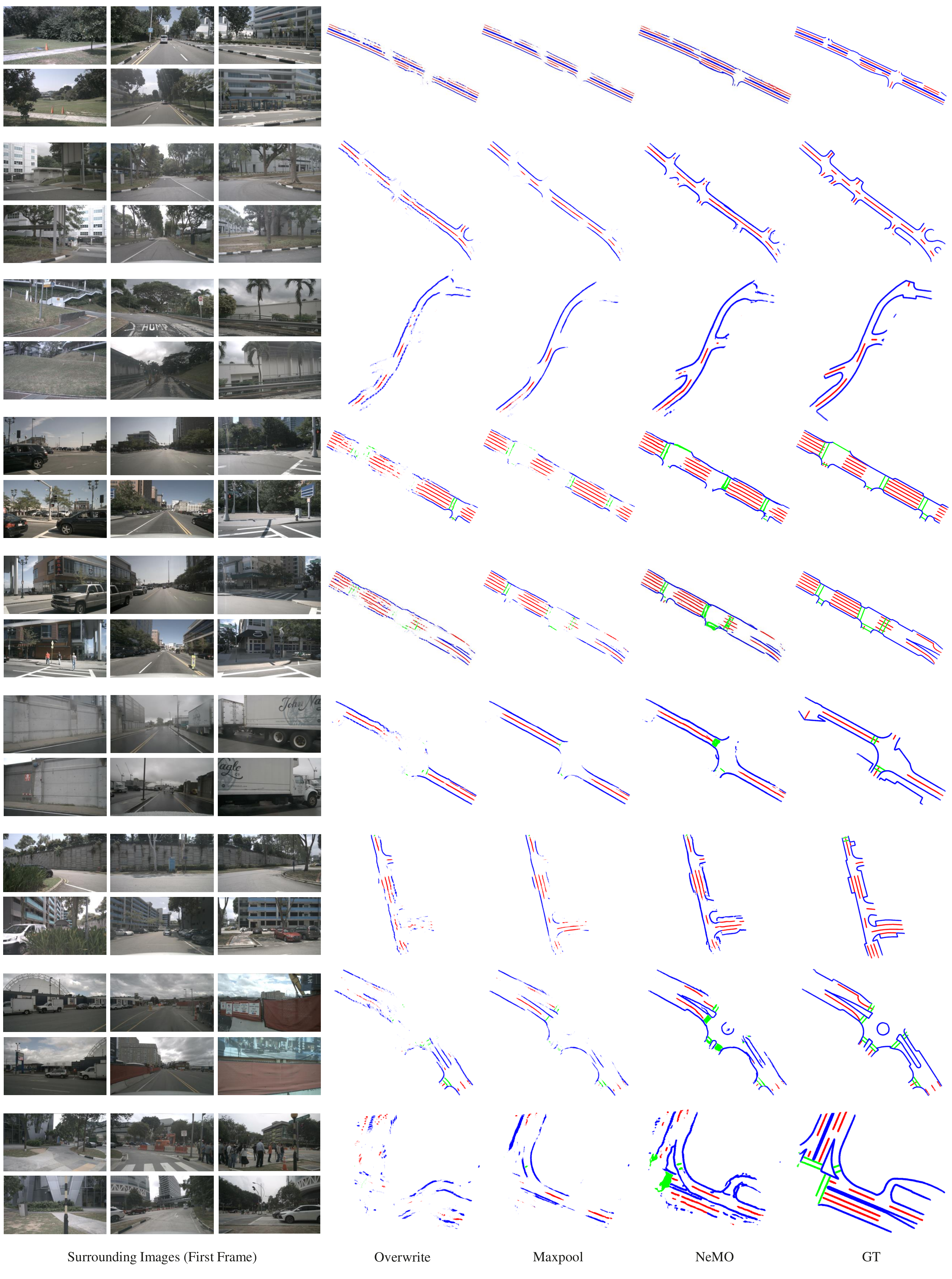}
	\caption{Visual comparison on NuScenes scene-level long-range map generation.}
	\label{fig:appn-nus-result}
\end{figure}

In Figure~\ref{fig:appn-nus-result} we provide more visualization results of scene-level long-range map on NuScenes dataset. It can be observed that NeMO surpasses two baselines by comprehensively capturing road elements while retaining more details, particularly in intersections. Notably, NeMO excels in capturing essential road structure information in complex scenes despite some noises in the generated map, as demonstrated in the final two cases in Figure~\ref{fig:appn-nus-result}. The overwriting and maxpooling techniques lead to comparatively poor performance primarily due to lower perception accuracy for elements at far distances compared to that of near ones in single-frame perception. As a result, during the time series update, there is a risk that higher accuracy perception results obtained at an earlier time may be overwritten and replaced by inferior results obtained later as the vehicle moves forward. Our method, however, addresses this issue by utilizing neural networks for effective information selection and update to better preserve accuracy over time. 

\begin{figure}[]
	\centering
	\begin{subfigure}[b]{0.7\textwidth}
		\centering
		\includegraphics[width=\linewidth]{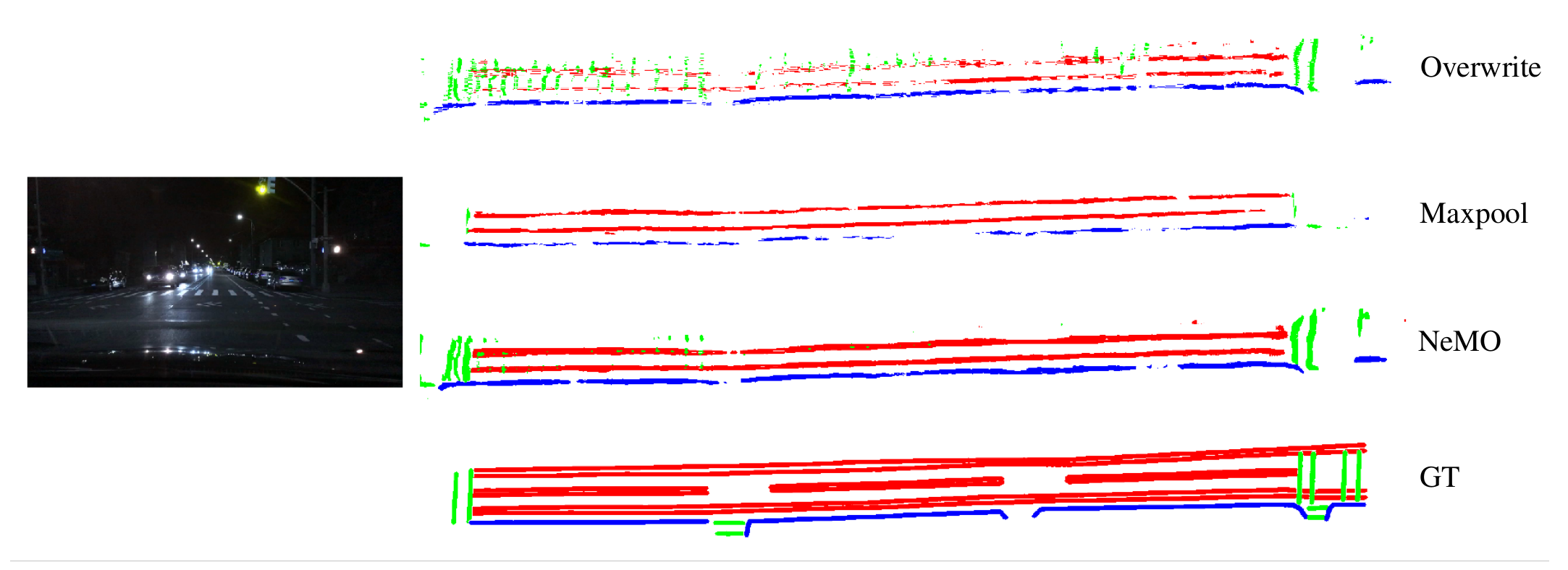}
	\end{subfigure}
	
	\begin{subfigure}[b]{0.7\textwidth}
		\centering
		\includegraphics[width=\linewidth]{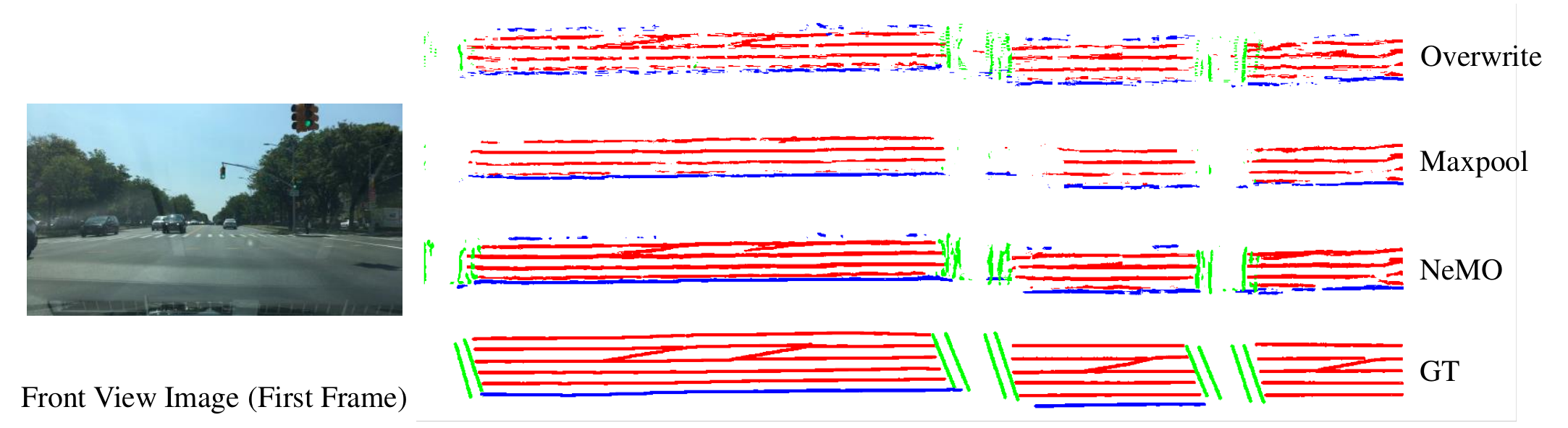}
	\end{subfigure}
	
	\caption{Visual comparison on BDD-Map scene-level long-range map generation (straight lanes).}
	\label{fig:appn-bdd-result0}
\end{figure}

\begin{figure}
	\centering
	\includegraphics[width=\linewidth]{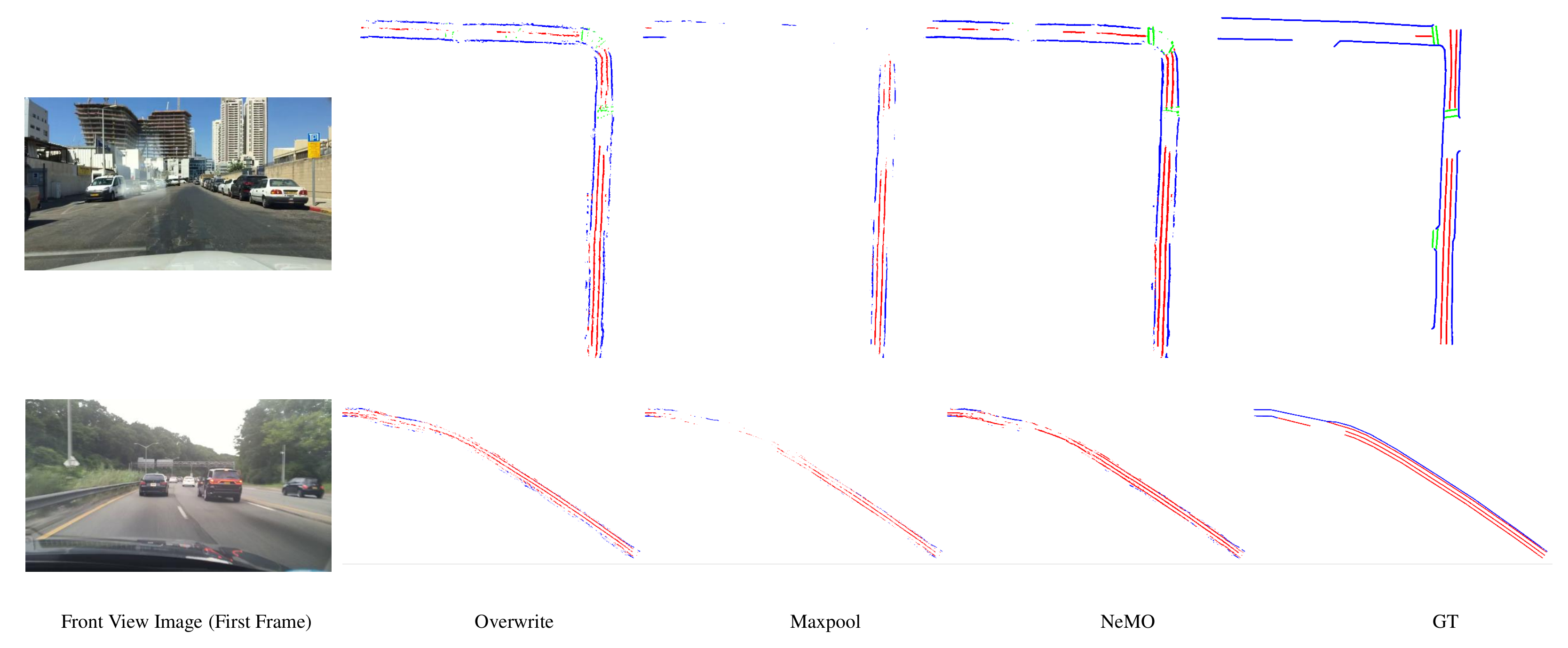}
	\caption{Visual comparison on BDD-Map scene-level long-range map generation (turning and curving).}
	\label{fig:appn-bdd-result1}
\end{figure}

Figure~\ref{fig:appn-bdd-result0} and Figure~\ref{fig:appn-bdd-result1} are visualizations of cases in BDD-Map dataset. In contrast to NuScenes, BDD-Map has more frames in scene clips and correspondingly longer-range local maps. NeMO paradigm proves highly effective in the creation of road structures in both straight road scenes (Figure~\ref{fig:appn-bdd-result0}) and turning scenes (Figure~\ref{fig:appn-bdd-result1}). These results indicate the continued viability of NeMO even in scenarios with less accurate sensor readings, underscoring its practical utility across a wide range of applications. Another observation is that in BDD-Map scenes, overwriting tends to outperform maxpooling approach. A possible explanation for this outcome is that the maxpooling approach that conserves maximum values can be more significantly influenced by pose noises. Specifically, multiple feature channels may retain large values in temporal update process, thus reducing the capacity of discrimination for key information.

\end{document}